\documentclass[letterpaper, 10 pt, conference]{ieeeconf}  

\IEEEoverridecommandlockouts                              

\usepackage{amsmath,amsfonts,amscd}
\usepackage{listings}
\usepackage[most]{tcolorbox}
\usepackage{color, colortbl}
\usepackage{float}
\usepackage{graphicx}
\usepackage{epstopdf}
\usepackage{verbatim}
\usepackage{tikz,times}
\usepackage{xcolor}
\usepackage{lipsum}
\usepackage{makecell}
\usepackage{booktabs}
\usepackage{algorithm}
\usepackage{algpseudocode}
\usepackage{tabularx}
\usepackage{cellspace}
\usepackage[Export]{adjustbox}
\usepackage{caption}
\usepackage{cuted}
\usepackage{pifont}
\usepackage{bbold}
\usepackage{dsfont}
\usepackage{multirow}

\newcommand{\cmark}{\ding{51}}%
\newcommand{\xmark}{\ding{55}}%

\newcommand{\OURS}{VLN-Zero}

\lstdefinelanguage{prompting}{
  morekeywords={Prompt, Response},
  sensitive=true,
  morecomment=[l]{--},
  morestring=[b]",
}
\lstdefinestyle{mylststyle}{
  basicstyle=\ttfamily\small,
  keywordstyle=\color{blue}\bfseries,
  commentstyle=\color{green!50!black}\itshape,
  stringstyle=\color{violet},
  showstringspaces=false,
  tabsize=2,
  breaklines=true,  
  columns=fullflexible, 
  keepspaces=true,
}

\tcbset{
  myboxstyle/.style={
    boxrule = 0pt,
    toprule = 4.5pt, 
    enhanced,
    fuzzy shadow = {0pt}{-2pt}{-0.5pt}{0.5pt}{black!35}, 
    fontupper = \ttfamily\small, 
    boxsep = 5pt, 
    left = 1pt, 
    right = 1pt, 
    top = 5pt, 
    bottom = 5pt, 
  }
}

\makeatletter
\let\NAT@parse\undefined
\makeatother
\usepackage{hyperref}
\hypersetup{
  colorlinks = true,
  linkcolor = red,
  anchorcolor = red,
  citecolor = cyan,
  urlcolor = cyan,
  pdftitle={MM3DGS SLAM}
}
\usepackage[capitalize]{cleveref}

\title{\LARGE \bf
\OURS: Rapid Exploration and Cache-Enabled Neurosymbolic Vision-Language Planning for Zero-Shot Transfer in Robot Navigation
}

\author{Neel P. Bhatt${}^{\dag}$, \IEEEmembership{Member, IEEE}, Yunhao Yang, \IEEEmembership{Member, IEEE}, Rohan Siva,\\ Pranay Samineni, Daniel Milan, Zhangyang Wang, and Ufuk Topcu, \IEEEmembership{Fellow, IEEE}
\thanks{All authors are with the University of Texas at Austin, Austin, TX, USA. ${}^\dag$Corresponding author; email: {\tt\small npbhatt@utexas.edu}. This work was supported by the Defense Advanced Research Projects Agency (DARPA) contract HR00112490431.}%
}

\begin{document}

\maketitle
\thispagestyle{empty}
\pagestyle{empty}

\begin{strip}
\begin{minipage}{\textwidth}\centering
\vspace{-25pt}
    \includegraphics[width=0.95\linewidth]{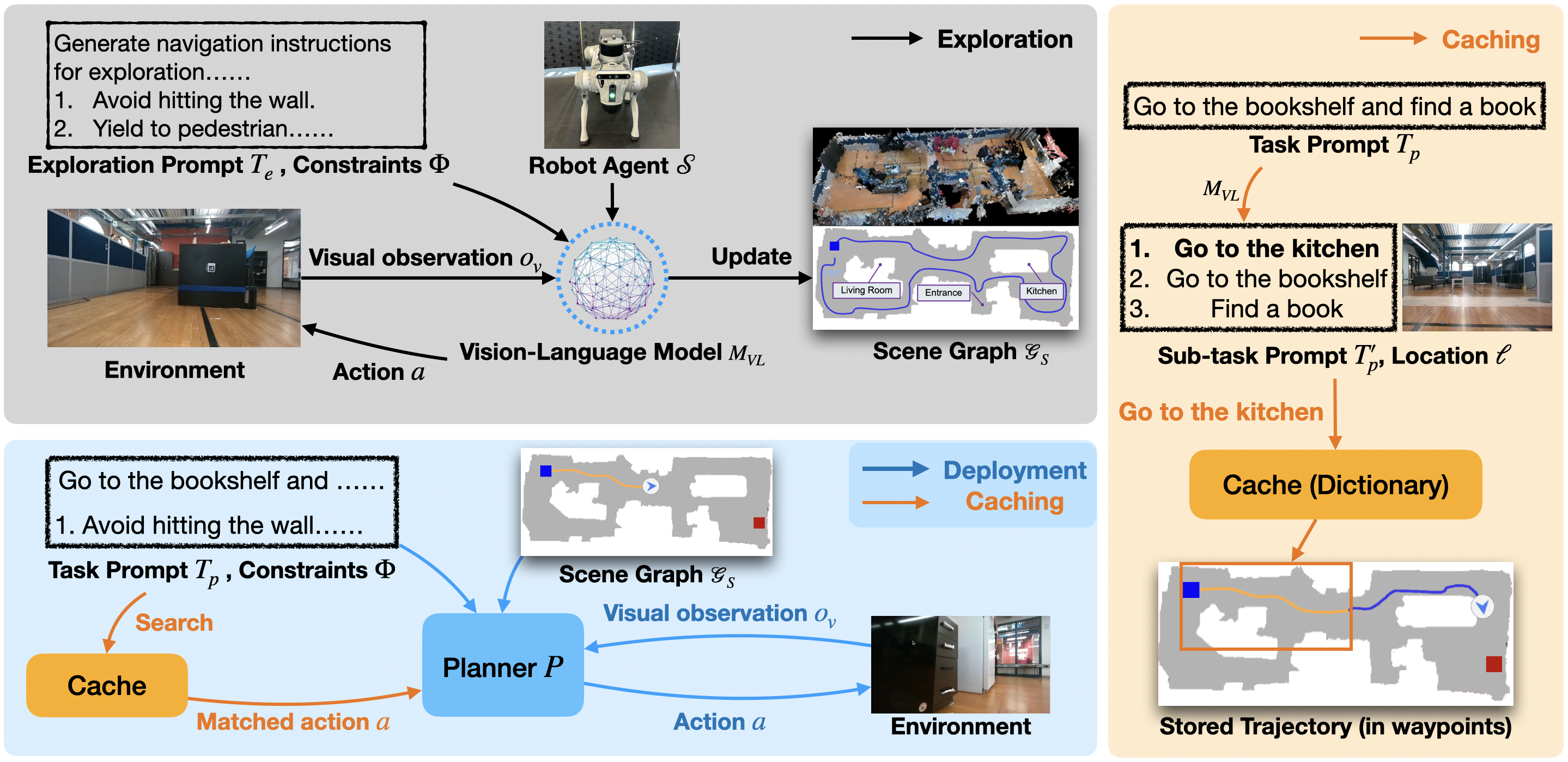}
    \captionof{figure}{\small Overview of \textbf{\OURS}. The framework consists of two main phases: exploration and deployment.
    In the \textbf{exploration phase} (gray), the VLM \(M_{VL}\) guides a robot agent \(\mathcal{S}\) to interact with the environment under user-specified constraints \(\Phi\), producing an action \(a\) at each step and ultimately yielding a scene graph \(G_S\) (top-down map). 
    In the \textbf{deployment phase} (blue), a planner \(P\) leverages the generated scene graph, visual observations, and constraints to generate constraint-satisfying actions for a task prompt \(T_p\). During execution, a caching module further accelerates execution by reusing previously validated trajectories via compositional task decomposition.}
    \label{fig: pipeline}
\end{minipage}
\end{strip}

\begin{abstract}


Rapid adaptation in unseen environments is essential for scalable real-world autonomy, yet existing approaches rely on exhaustive exploration or rigid navigation policies that fail to generalize. We present \OURS, a two-phase vision-language navigation framework that leverages vision-language models to efficiently construct symbolic scene graphs and enable zero-shot neurosymbolic navigation. In the exploration phase, structured prompts guide VLM-based search toward informative and diverse trajectories, yielding compact scene graph representations. In the deployment phase, a neurosymbolic planner reasons over the scene graph and environmental observations to generate executable plans, while a cache-enabled execution module accelerates adaptation by reusing previously computed task–location trajectories. By combining rapid exploration, symbolic reasoning, and cache-enabled execution, the proposed framework overcomes the computational inefficiency and poor generalization of prior vision-language navigation methods, enabling robust and scalable decision-making in unseen environments. \OURS \ achieves 2x higher success rate compared to state-of-the-art zero-shot models, outperforms most fine-tuned baselines, and reaches goal locations in half the time with 55\% fewer VLM calls on average compared to state-of-the-art models across diverse environments.{\let\thefootnote\relax\footnote{The full codebase, datasets, and videos for \OURS \ are available at \url{https://vln-zero.github.io/}.}}


\end{abstract}

\section{INTRODUCTION}

Deploying autonomous agents in new environments remains a fundamental challenge: policies trained in one setting often fail in another due to novel layouts, obstacles, or constraints, and consequently require fine-tuning or multi-shot inference. 
For example, a robot trained to navigate one office building may struggle in a different building unless retrained, a process that is both slow and impractical for rapid deployment.

The central problem is twofold: (i) How can an agent efficiently construct a representation of an unseen environment, such as a scene graph, without exhaustive search? 
(ii) Given this representation, how can the agent \textit{efficiently} generate constraint-satisfying plans in real time \textit{without fine-tuning} or \textit{multi-shot} inference?

Existing approaches, ranging from frontier-based exploration to reinforcement learning with fixed policies, struggle with either computational inefficiency, lack of generalization, or both. Vision-language navigation models (VLNs) offer promise; however, current approaches suffer from slow, exhaustive exploration, weak task decomposition, and high training and query cost. 

We argue that addressing these limitations requires rethinking the interaction between perception, symbolic reasoning, and policy adaptation. Specifically, agents must be able to (i) rapidly acquire symbolic representations of their environment to minimize exploration cost, and (ii) leverage these representations for efficient navigation in new environments without retraining or extensive fine-tuning.

To address these challenges, we introduce \OURS , a two-phase \textit{zero-shot} framework that combines vision-language model (VLM) guided exploration with neurosymbolic navigation. In the exploration phase, the agent interacts with the environment using structured and compositional task prompts, guiding exploration toward informative and diverse trajectories to construct a compact scene graph with semantic area labels. In the deployment phase, a neurosymbolic planner reasons over this scene graph, environmental observations to generate executable plans, eliminating reliance on fixed policies. To further improve scalability, we propose a cache-enabled execution procedure that stores previously computed task–location trajectories for reuse, accelerating both exploration and deployment.

In summary, \OURS \ offers three key contributions:
\begin{itemize}
    \item \textbf{VLM-guided rapid exploration:} We design structured, compositional prompts that steer a VLN agent to propose exploration actions while incrementally constructing compact symbolic scene graphs. This enables coverage of novel environments within a time- and compute-constrained exploration budget while avoiding unsafe behaviors.  

    \item \textbf{Zero-shot neurosymbolic navigation:} We introduce a planner that reasons jointly over 
    scene graphs, task prompts, and real-time observations, transforming free-form natural language 
    instructions into constraint-satisfying action sequences without fine-tuning or multi-shot inference.  

    \item \textbf{Cache-enabled execution for fast adaptation:} We develop a trajectory-level caching mechanism 
    that stores validated task--location pairs, allowing the system to reuse previously computed plans. 
    This reduces redundant VLM queries to minimize execution time, cost, and compute demands which accelerates real-world deployment.
\end{itemize}

\section{RELATED WORKS}
Zero-shot VLN approaches such as CoW-OWL, AO-Planner, A²Nav, NavGPT-CE, OpenNav, InstructNav, and CA-Nav leverage foundation models, LLM reasoning, affordances, and structured sub-instruction decomposition to achieve relatively strong performance~\cite{gadre2022cowspasturebaselinesbenchmarks, chen2024affordancesorientedplanningusingfoundation, chen2023a2navactionawarezeroshotrobot, zhou2023navgpt, qiao2025opennavexploringzeroshotvisionandlanguage, long2024instructnavzeroshotgenericinstruction, chen2025constraintawarezeroshotvisionlanguagenavigation}. While these methods are low-cost and highly flexible, performance still generally lags behind fine-tuned models \cite{wang2024planning}.

Fine-tuned, task-specific vision-language models have demonstrated strong performance in embodied navigation. Methods such as GridMM, LAW, DREAMWALKER, and GridMM are able to utilize large-scale training to encode spatio-temporal context, structured maps, or abstract world models, enabling agents to follow instructions effectively in both discrete and continuous 3D environments~\cite{krantz2022sim2sim, wang2023gridmm, Wang_2023_ICCV, chen2020topologicalplanningtransformersvisionandlanguage, raychaudhuri2021language, krantz_vlnce_2020}. Furthermore, NaVid~\cite{2402.15852} and NaVILA~\cite{cheng2024navila} demonstrate that with extensive training, agents can achieve high success rates in indoor navigation using only POV camera inputs.

\paragraph{Exploration and Environment Representation}  
Efficient exploration is fundamental to autonomous decision-making. Classical methods in frontier-based exploration and occupancy-grid mapping aim to reduce uncertainty through exhaustive coverage~\cite{yamauchi1997frontier,thrun2005probabilistic}, but they are often computationally expensive and scale poorly in large or dynamic environments. More recent approaches leverage semantic representations such as scene graphs and object-centric maps to capture relational structure~\cite{wu2017ai2thor,krishna2017visual}, enabling higher-level reasoning about tasks. However, these methods typically rely on either hand-engineered heuristics or environment-specific priors, limiting their applicability to unseen domains.  

\paragraph{Navigation and Adaptation in New Environments}  
Traditional navigation approaches, ranging from symbolic planners such as PDDL-based systems~\cite{helmert2006fast} to modern hierarchical reinforcement learning~\cite{barto2003recent}, struggle with generalization when deployed in novel environments. Symbolic methods offer interpretability but often lack robustness to perceptual noise, while learned policies achieve strong performance in-distribution but require costly retraining for new tasks. Neurosymbolic approaches~\cite{garcez2019neurosymbolic,belle2021neurosymbolic, nsaipnas} have emerged as a promising middle ground by combining structured reasoning with data-driven learning, yet they typically assume fixed task specifications and do not explicitly address rapid adaptation with minimal refinement.  

\paragraph{Shortcomings with Current Solutions}  
Current VLM-based navigation frameworks suffer from three key shortcomings: (i) exploration remains inefficient, often requiring dense interaction with the environment \cite{bhattknow, hua2021learning, yang2024fine}; (ii) hierarchical task decomposition is either hand-specified or absent, limiting scalability \cite{li2021safe, liang2024skilldiffuser, yang2024joint}; and (iii) computational cost is high, restricting real-time applicability \cite{nygaard2021environmental, hu2025flare, jeong2024survey}.  
  
Our framework addresses these gaps by combining zero-shot VLM-guided exploration with scene-graph-based symbolic representations and cache-enabled neurosymbolic navigation. Unlike prior approaches that rely on highly fine-tuned models, exhaustive search, or fixed policies, our method enables efficient representation learning during exploration and flexible adaptation at deployment. This integration allows agents to rapidly and successfully generalize navigation policies to unseen environments while maintaining computational efficiency.

\section{PROBLEM FORMULATION}

\OURS \ leverages a VLM to enable \emph{rapid exploration and adaptation} without the need to fine-tune or perform multi-shot inference. The VLM is informed via textual constraints and guides exploration to build a scene graph of the environment. During deployment, this scene graph, visual observations, and cached policies are consumed by the VLM to enable efficient execution while enforcing user-specified constraints.

Formally, we consider an autonomous agent \(\mathcal{S}\) equipped with sensors such as a camera and an inertial measurement unit (IMU) along with compute offering API access to call a VLM. The agent operates under a set \(\Phi\) of user-provided constraints, expressed in natural language (e.g., ``Do not collide with pedestrians''), which must hold during both exploration and execution. From the camera, the agent obtains first-person images as visual observations \(o\). A pre-trained VLM \(M_{VL}\) serves as a reasoning engine with both image-understanding and text-reasoning capabilities.  

The input to \(M_{VL}\) consists of the tuple \((\mathcal{S}, \Phi, o)\). A call to \(M_{VL}\) yields the outputs:  
\begin{itemize}
    \item an action \(a\) to be executed by the agent, and  
    \item a \emph{scene graph} \(G_S\) during exploration, represented as a \emph{top-down map of the environment}, capturing traversable areas, obstacles, and landmarks used for downstream navigation.  
\end{itemize}

We formalize two central problems for this work.  

\paragraph{Problem 1 - Rapid Exploration}  
We want to design a prompting strategy for \(M_{VL}\) that enables the autonomous agent to efficiently explore a novel environment and facilitate construction of a top-down scene graph \(G_S\) as a byproduct of the navigation actions taken. The exploration process should be completed within a limited time span of \emph{1 hour}, producing a scene graph sufficient to support navigation. 

\paragraph{Problem 2 - Deployment} 
Construct a planner \(P\) such that
\[
P(\Phi, o, T_p, G_S) = a ,
\]
where \(T_p\) is a textual task prompt describing the agent’s expected navigation goal. The planner generates an action \(a\) that should (i) satisfy \(\Phi\), and (ii) contribute to efficiently achieving the navigation objective in \(T_p\). Importantly, the planner should adapt to the new environment by reasoning over the scene graph \(G_S\) obtained during exploration, \emph{without requiring fine-tuning}.

\section{\OURS: ZERO-SHOT VLM NAVIGATION}
We develop a framework, \OURS, that consists of two phases: an \emph{exploration phase}, where the autonomous agent leverages a VLM to rapidly explore a new environment and construct a top-down scene graph, and a \emph{deployment phase}, where the planner uses this scene graph to adapt to new tasks under user-specified constraints without fine-tuning. The deployment phase further integrates a high-level landmark-to-landmark cache that reuses previously validated plans to accelerate execution.

\begin{figure}[t]
    \centering
\begin{tcolorbox}[
  myboxstyle
]
\begin{lstlisting}[
language=prompting,
style=mylststyle,
]
Prompt: Generate simple step-by-step navigation instructions for a robot exploring an unknown environment. The robot can only perform the following actions: move forward, turn left, turn right, and stop.

Constraints:
- Use only the allowed actions.
- Return a single action.
- Stop when exploration of the visible environment is complete or when further movement is unsafe.
- Avoid visiting explored areas

<Scene Graph>

<Visual Observation>

Response: <One of the four actions> 
\end{lstlisting}
\end{tcolorbox}
\caption{Prompt $T_e$ used for guiding \(M_{VL}\) during the exploration phase.}
\label{fig:exploration-prompt}
\end{figure}

\begin{algorithm}[tb]
\caption{\OURS: VLM-Guided Exploration}
\label{alg:exploration}
\begin{algorithmic}[1]
\Require Agent $\mathcal{S}$, constraints $\Phi$, VLM $M_{VL}$
\Ensure Scene graph $G_S$
\State Initialize empty scene graph $G_S \gets \emptyset$
\State $t \gets 0$
\While{$t < \text{exploration\_time\_limit}$}
    \State Observe environment: $o \gets \text{sensor\_data()}$
    \State $(a, G_S') \gets M_{VL}(\mathcal{S}, \Phi, o)$
    \State Execute action: $\mathcal{S}.\text{execute}(a)$
    \State Update scene graph: $G_S \gets G_S \cup G_S'$
    \State $t \gets t + 1$
\EndWhile
\State \Return $G_S$
\end{algorithmic}
\end{algorithm}

\subsection{Rapid Exploration}
The goal of the exploration phase is to enable the autonomous agent \(\mathcal{S}\) to rapidly adapt to an unseen environment without retraining. Instead of relying on traditional pre-trained planner that may not generalize, we leverage the multimodal reasoning ability of the VLM \(M_{VL}\) to jointly interpret user constraints \(\Phi\) and visual observations \(o\). The VLM produces exploratory high-level actions, e.g., move straight or turn right, that safely guide the robot through the environment used to incrementally construct an interpretable scene graph \(G_S\) in the form of a top-down map using odometry from an IMU.

At each step of the exploration process, the autonomous agent (i.e., robot) obtains a new observation \(o\) through its visual sensors. The tuple \((\mathcal{S}, \Phi, o)\) is sent to \(M_{VL}\) via an API call. 
The model outputs an action \(a\), which is executed by the robot, and a partial update \(G_S'\) is made to the scene graph.
The global scene graph \(G_S\) is updated constantly by merging \(G_S'\) into the existing representation. During this process, we group objects that are semantically relevant to landmarks such as ``living room" or ``kitchen" to append semantic tags to the scene graph. By prompting the VLM with \(\Phi\), unsafe behaviors (e.g., entering restricted areas or colliding with obstacles) are avoided during exploration. We present the input prompt to the VLM that guides the exploration, as shown in Fig. \ref{fig:exploration-prompt}.

We present this exploration process in Algorithm~\ref{alg:exploration}. 
This exploration loop continues until either the exploration time budget is exhausted or sufficient coverage of the environment has been achieved.
To detect whether the exploration phase achieves sufficient coverage, we monitor the top-down scene graph for two structural indicators: (i) the loop closure of the periphery, indicating that the robot has traced the boundaries of the accessible region, and (ii) the absence of large unexplored empty areas within the mapped region. We present an example of the constructed scene graph with the exploration route in Fig. \ref{fig: pipeline}.
Once both conditions are satisfied, we consider the environment sufficiently explored, and the final scene graph \(G_S\) is ready for deployment.

\begin{figure}[t]
    \centering
\begin{tcolorbox}[
  myboxstyle
]
\begin{lstlisting}[
language=prompting,
style=mylststyle,
]
Prompt: Move from your start location to a goal location using the provided top-down scene graph and camera view.

<Constraints>

<Scene Graph>

- SQUARE: Your starting position.
- BLUE ARROW: Your current position & heading.
- BLUE LINE: Your trajectory so far.
- GRAY AREAS: Navigable floor where you can walk.
- WHITE AREAS: Obstacles or walls you cannot walk through.

<Visual Observation>

NAVIGATION RULES:
- Use a top-down scene graph to determine the direction to move in.
......

Response: <One of the four actions> 
\end{lstlisting}
\end{tcolorbox}
\caption{Prompt $T_p$ used for guiding the planner \(P\) during the deployment phase. Text in brackets ($<>$) are parameterized, while the others are fixed meta prompts. We present a scene graph used in this prompt in Fig. \ref{fig: topdownmap-demo}.}
\label{lst:deployment-prompt}
\end{figure}

\begin{algorithm}[tb]
\caption{\OURS: Cached-Enabled Navigation}
\label{alg:planning}
\begin{algorithmic}[1]
\Require Prompt $T_p$, location $\ell$, scene graph $G_S$, cache $\mathcal{C}$
\Ensure Action $a$

\If{$(T_p) \in \mathcal{C}$} \Comment{Check task-level cache}
    \State $\tau \gets \mathcal{C}[(T_p)]$
    \State $a \gets \text{next\_action}(\tau, \ell)$

\ElsIf{$(T_p', \ell) \in \mathcal{C}$} 
    \Comment{Check subtask/location cache, where $T_p'$ is the subtask/location-specific query}
    \State $\tau \gets \mathcal{C}[(T_p', \ell)]$
    \State $a \gets \text{next\_action}(\tau)$

\Else
    \Comment{Fall back to navigation}
    \State $a \gets P(\Phi, o, T_p, G_S)$
    \State $\tau \gets \text{compute\_trajectory}(a, T_p, G_S)$
    \State $\mathcal{C}[(T_p, \ell)] \gets \tau$ 
    \Comment{Store subtask-level plan}
    \State $\mathcal{C}[(T_p)] \gets \text{merge}(\mathcal{C}[(T_p)], \tau)$ 
    \Comment{Update task plan}
\EndIf

\State \Return $a$
\end{algorithmic}
\end{algorithm}

In practice, we enforce a 1 hour limit, which balances the need for rapid deployment with the completeness of the resulting scene graph. 
The final scene graph \(G_S\) serves as the structured foundation for downstream navigation in the deployment phase.

\subsection{Cache-Enabled Neurosymbolic Deployment}

\paragraph{Environment Adaptation Using a Scene Graph} 
In the deployment phase, the planner $P$ leverages the scene graph \(G_S\) constructed during exploration to adapt to the new environment without any additional training. 
Unlike policy-based methods that require re-training, the planner adapts to the new environment by reasoning directly over the scene graph (i.e., top-down map). 
The planner \(P\) receives the task prompt \(T_p\), constraints \(\Phi\), real-time observations \(o\) that include a first-person-view image and a top-down scene graph \(G_S\) as inputs. We present a sample prompt in Fig. \ref{lst:deployment-prompt}.
The planner, \(P(\Phi, o, T_p, G_S)\), computes a constraint-compliant action that contributes to accomplishing the navigation goal by reasoning over the scene graph. For example, given the task prompt ``deliver an item to the kitchen,'' the planner uses \(G_S\) to identify feasible trajectories, ensures that non-traversable areas are excluded, and generates an action \(a\) aligned with the task objective.

\paragraph{Task Execution with Caching}  
To accelerate real-time navigation, we introduce a caching module that stores previously validated navigation trajectories. Given, VLMs do not have an explicit internalization of memory, intuitively, caching helps with both efficiency and task success probability. 
Formally, the cache $\mathcal{C}$ is maintained as a dictionary of the form
\[
\mathcal{C} = \{(T_p, \text{current location } \ell) : \text{trajectory}\},
\]
where each entry associates a task prompt and the agent’s current position with a feasible trajectory that satisfies the constraints \(\Phi\), a \textit{trajectory} is a set of consecutive points in the scene graph representing the relative positions in the environment. We present an example of a cached trajectory in Fig. \ref{fig: pipeline}.

To further accelerate execution and improve reuse across tasks, we employ a \textbf{hierarchical caching} strategy. 
Instead of storing only complete trajectories for specific task--location pairs, the cache \(\mathcal{C}\) maintains trajectories at multiple granularities: task-level trajectories \((T_p)\), subtask or location-specific trajectories \((T_p, \ell)\), and reusable fragments such as room-to-room or room-to-object transitions. For example, a room-to-room trajectory will be stored as 

\noindent \textit{$\{$(``go to the bedroom'', $\ell=$location of the main entrance): a trajectory from the main entrance to the entrance of the bedroom$\}$.}

And a sample object-to-object transition is

\noindent \textit{$\{$(``go to bed'', $\ell=$entrance of the bedroom): a trajectory from the bedroom entrance to the bed$\}$.}

This hierarchical structure enables the planner to compose cached sub-trajectories to construct feasible trajectories for new tasks, even when the exact query has not been encountered before.  

During execution, the planner first checks whether the cache already contains a task-level trajectory for \(T_p\). 
If available, the next action is directly extracted from this stored plan instead of calling the VLM. 
If not, the planner first decomposes the task prompt into a set of subtask- or location-specific prompts. 
In practice, we use a VLM as our planner backbone and query it for task decomposition.
Then, the planner queries the subtask or location-level cache, retrieving smaller route segments that can be reused. 
When neither level of cache contains a matching entry, the planner falls back to computing a new trajectory by reasoning over the scene graph \(G_S\). 

We illustrate the overall process in Algorithm~\ref{alg:planning}, which demonstrates how hierarchical caching balances efficiency and adaptability. In summary, our deployment phase highlights the two complementary modes of operation: 
(i) fast retrieval of validated trajectories when available, and (ii) adaptive reasoning over the scene graph when new task queries are encountered. 
This dual mechanism ensures that the planner can adapt to novel navigation goals in unfamiliar environments while maintaining efficiency and adaptability.  


\section{Results}


\begin{table*}[t]
    \centering
    \caption{Cross-dataset performance on the R2R (Val-Unseen) and RxR (Val-Unseen) benchmarks. We group prior works into non-zero-shot and zero-shot frameworks. \OURS \ outperforms all zero-shot frameworks across all metrics in both R2R and RxR. Compared to fine-tuned baselines, \OURS \ achieves on-par performance on both R2R and RxR datasets.
    }
    \vspace{-4pt}
    \begin{tabular}{c|c|cccc|cccc|cccc} 
    \toprule
    \multirow{2}{*}{Method} & \multirow{2}{*}{Zero-Shot} &
    \multicolumn{4}{c|}{Observation} &
    \multicolumn{4}{c|}{R2R Val-Unseen} &
    \multicolumn{4}{c}{RxR Val-Unseen} \\[0.8ex] 
    \cline{3-14}
    & & \rule{0pt}{2.75ex} S.RGB & Pano. & Depth & Odo. & NE ↓ & OS ↑ & SR ↑ & SPL ↑ & NE ↓ & SR ↑ & SPL ↑ & OS ↑ \\
    \midrule
    Sim2Sim \cite{krantz2022sim2sim} & \xmark &  & \checkmark & \checkmark & \checkmark & 6.07 & 52.0 & 43.0 & 36.0 & -- & -- & -- & -- \\
    GridMM \cite{wang2023gridmm} & \xmark &  & \checkmark & \checkmark & \checkmark & 5.59 & 56.0 & 47.0 & 41.0 & -- & -- & -- & -- \\
    DreamWalker \cite{Wang_2023_ICCV} & \xmark &  & \checkmark &  & \checkmark & 5.30 & 64.0 & 51.0 & 46.0 & -- & -- & -- & -- \\
    AG-CMTP \cite{chen2020topologicalplanningtransformersvisionandlanguage} & \xmark &  & \checkmark & \checkmark & \checkmark & 7.90 & 39.0 & 26.0 & 22.0 & -- & -- & -- & -- \\
    R2R-CMTP \cite{chen2020topologicalplanningtransformersvisionandlanguage} & \xmark &  & \checkmark & \checkmark & \checkmark & 7.93 & 26.0 & 24.0 & 20.0 & 10.90 & 8.0 & 8.0 & 38.0 \\
    LAW \cite{raychaudhuri2021language} & \xmark & \checkmark & \checkmark & \checkmark & \checkmark & 7.90 & 26.0 & 24.0 & 20.0 & 10.87 & 8.0 & 8.0 & 21.0 \\
    Seq2Seq \cite{krantz_vlnce_2020} & \xmark & \checkmark &  & \checkmark &  & 7.77 & 30.0 & 25.0 & 20.0 & 12.10 & 13.9 & 11.9 & 31.0 \\
    RGB-Seq2Seq \cite{krantz_vlnce_2020} & \xmark & \checkmark &  & \checkmark &  & 10.80 & 12.0 & 10.0 & 8.0 & 11.2 & 0.0 & 0.0 & 12.2 \\
    Navid \cite{2402.15852} & \xmark & \checkmark &  &  &  & 5.47 & 49.0 & 37.0 & 35.0 & 8.41 & 23.8 & 21.2 & 34.5 \\
    NaVILA \cite{cheng2024navila} & \xmark & \checkmark &  &  &  & 5.22 & 62.5 & 54.0 & 49.0 & 8.78 & 34.3 & 28.2 & 46.8 \\
    \midrule
    CoW-OWL \cite{gadre2022cowspasturebaselinesbenchmarks} & \cmark & \checkmark &  & \checkmark &  & 8.72 & 5.9 & 3.4 & 1.6 & 8.93 & 14.4 & 9.2 & 25.0 \\
    CoW-CLIP-Grad \cite{gadre2022cowspasturebaselinesbenchmarks} & \cmark & \checkmark &  & \checkmark &  & 8.68 & 3.4 & 1.8 & 1.3 & -- & -- & -- & -- \\
    AO-Planner \cite{chen2024affordancesorientedplanningusingfoundation} & \cmark & & \checkmark & \checkmark &  & 6.95 & 38.3 & 25.5 & 16.6 & 10.75 & 22.4 & 15.1 & -- \\
    A$^2$Nav \cite{chen2023a2navactionawarezeroshotrobot} & \cmark & \checkmark &  &  &  & -- & -- & 22.6 & 11.1 & -- & 16.8 & 6.3 & -- \\
    NavGPT-CE \cite{zhou2023navgpt} & \cmark & \checkmark &  & \checkmark &  & 8.37 & 26.9 & 16.3 & 10.2 & -- & -- & -- & -- \\
    OpenNav  \cite{qiao2025opennavexploringzeroshotvisionandlanguage} & \cmark & \checkmark & \checkmark  & \checkmark &  & 7.25 & 23.0 & 16.0 & 12.9 & -- & -- & -- & -- \\
    InstructNav \cite{long2024instructnavzeroshotgenericinstruction} & \cmark & \checkmark & \checkmark  & \checkmark  &  & 9.20 & 47.0 & 17.0 & 11.0 & -- & -- & -- & -- \\
    CA-Nav \cite{chen2025constraintawarezeroshotvisionlanguagenavigation} & \cmark & \checkmark & \checkmark & \checkmark & \checkmark & 7.58 & 48.0 & 25.3 & 10.8 & 10.37 & 19.0 & 6.0 & -- \\
    \textbf{\OURS \ (Ours)}  & \cmark & \checkmark &  &  & \checkmark & \textbf{5.97} & \textbf{51.6} & \textbf{42.4} & \textbf{26.3} & 9.13 & \textbf{30.8} & \textbf{19.0} & \textbf{37.5} \\
    \bottomrule
    \end{tabular}
    \label{tab:methodresults}
\end{table*}

\textbf{Setup:}
To evaluate \OURS, we use the widely popular Habitat Simulator \cite{savva2019habitatplatformembodiedai}.  Habitat-Sim is a 3D simulation platform that supports a variety of different datasets and is particularly useful for evaluating and developing embodied agents. We build our simulation environment based on the VLN-CE environment \cite{2402.15852} and evaluate our agent on the R2R \cite{anderson2018visionandlanguagenavigationinterpretingvisuallygrounded} and RxR \cite{ku2020room} room-to-room datasets. The val-unseen splits of these datasets are standard benchmarks for VLN with 1,839 and 1,517 episodes, respectively \cite{Gu_2022}. Each \emph{episode} includes a set of navigable instructions for a task, which is associated with a start and goal location. However, note that our method does not rely on these instructions. Our framework uses API calls to query the GPT-4.1 or model with our prompt, along with an updated RGB camera view and top-down scene graph map from the Habitat-Sim agent.

\textbf{Evaluation Metrics:} In our experiments, we report standard VLN evaluation metrics, including navigation error (NE), oracle success rate (OS), success rate (SR) and success-weighted path length (SPL). An episode is considered a success if the agent stops within 3 meters of the goal in the simulation in both RxR and R2R datasets. 

\textbf{Baselines:} We compare against 8 zero-shot and 10 fine-tuned frameworks as baselines. We list four state-of-the-art (SOTA) zero-shot and finetuned models below:
\begin{itemize}
    \item \textbf{CA-Nav}\cite{chen2025constraintawarezeroshotvisionlanguagenavigation} is a zero-shot model that guides navigation by breaking instructions into sub-tasks and enforcing constraints to generate actions.

    \item \textbf{AO-Planner}\cite{chen2024affordancesorientedplanningusingfoundation} is a zero-shot model that connects high-level instruction following with low-level motion control by leveraging visual affordances to plan and execute navigation paths.

    \item \textbf{NaVid}\cite{2402.15852} is a fine-tuned video-based vision-language model that navigates unseen environments by following language instructions and utilizing image history for decision-making.

    \item \textbf{NaVILA}\cite{cheng2024navila} is a finetuned vision-language-action framework for legged robots that follows language instructions by generating spatial actions, which are then executed through a low-level locomotion policy.
\end{itemize}

\subsection{Simulation Results on R2R and RxR Datasets}

We first evaluate \OURS \ on the VLN-CE R2R Val-Unseen dataset \cite{anderson2018visionandlanguagenavigationinterpretingvisuallygrounded}. Evaluation results of our model on the VLN-CE benchmarks are presented in \cref{tab:methodresults} and compared to existing navigation frameworks. The table outlines input modalities required for each method, where \OURS \ only requires  RGB images and odometry. 
As shown in \cref{tab:methodresults}, \OURS \ achieves an SR of 42.4\%, exceeding the best zero-shot baseline by over 17\% (CA-Nav at 25.3\%). We also outperform in OS (51.6\% vs 48.0\% by CA-Nav). Our SPL also improves to 26.3\%, more than doubling the SPL of Open-Nav (12.9\%) and surpassing all other zero-shot methods. Being a zero-shot framework, \OURS \ also outperforms most fine-tuned SOTA models. 


We then evaluate on the VLN-CE RxR Val-Unseen dataset \cite{ku2020room}. Evaluation of \OURS \ can be compared with existing zero-shot navigation frameworks in \cref{tab:methodresults}. Our method outperforms the second-best zero-shot method, AO Planner, by 8.4\%. \OURS's OS additionally surpasses all zero-shot results at 37.5\%. We also exceed AO-Planner in SPL by 3.9\%. 

Across both datasets, we show that \OURS \ \textbf{outperforms existing SOTA zero-shot frameworks}. Compared with the approaches that require training or fine-tuning, \OURS \ also achieves \textbf{at par performance with the fine-tuned SOTA models, without reliance on fine-tuning}. 
Evaluations across both datasets demonstrate that \OURS\ reduces the training time and computational overhead while retaining strong generalization and scalability.

\begin{figure*}[t]
\setlength\tabcolsep{0.5pt}
\adjustboxset{width=\linewidth,valign=c}
\centering
\begin{tabular}{c  c c}
    & \OURS\ (GPT-5) & \OURS\ (GPT-4.1)\\

    \rotatebox[origin=c]{90}{Episode 331} &
    \includegraphics[width=0.4\linewidth]{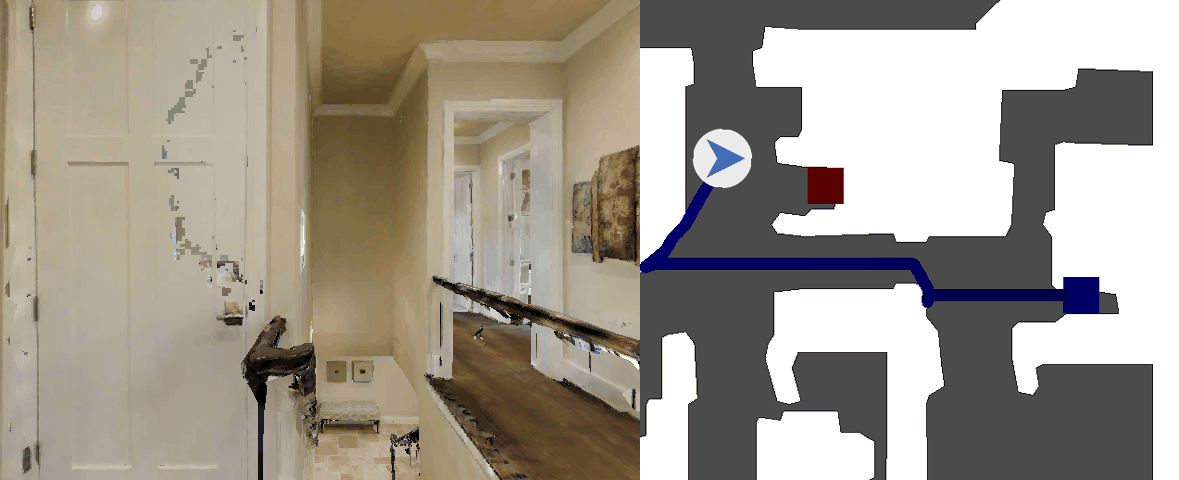} &
    \includegraphics[width=0.4\linewidth]{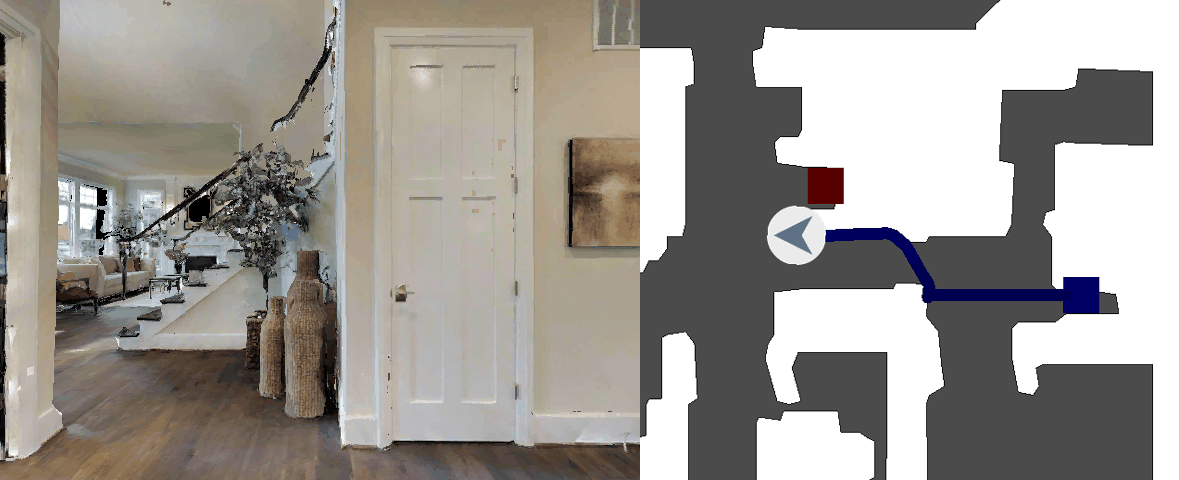} \\

    \rotatebox[origin=c]{90}{Episode 800} &
    \includegraphics[width=0.4\linewidth]{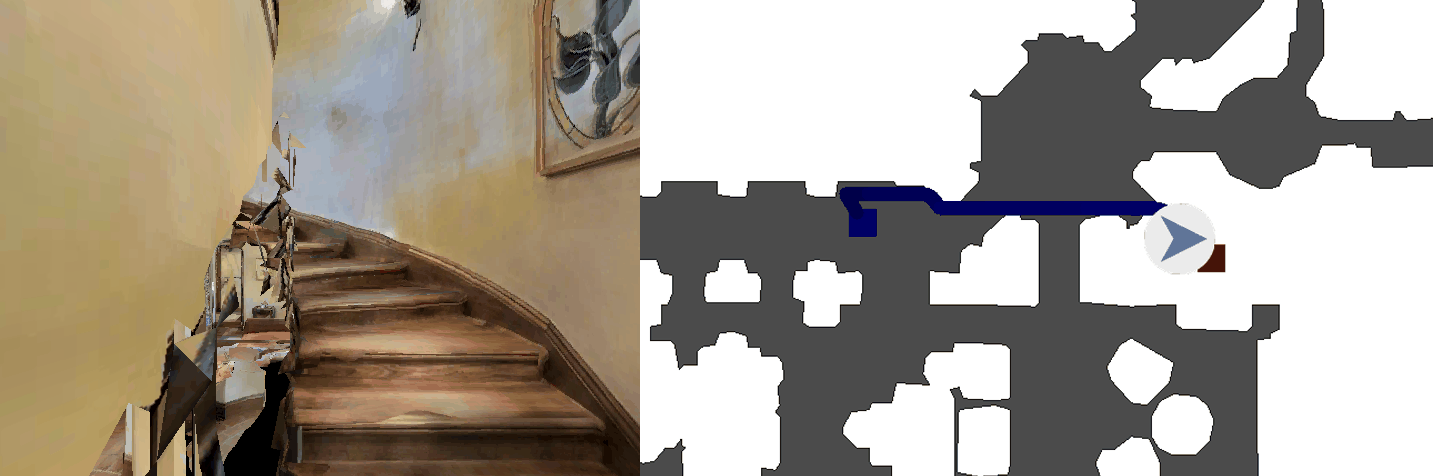} &
    \includegraphics[width=0.4\linewidth]{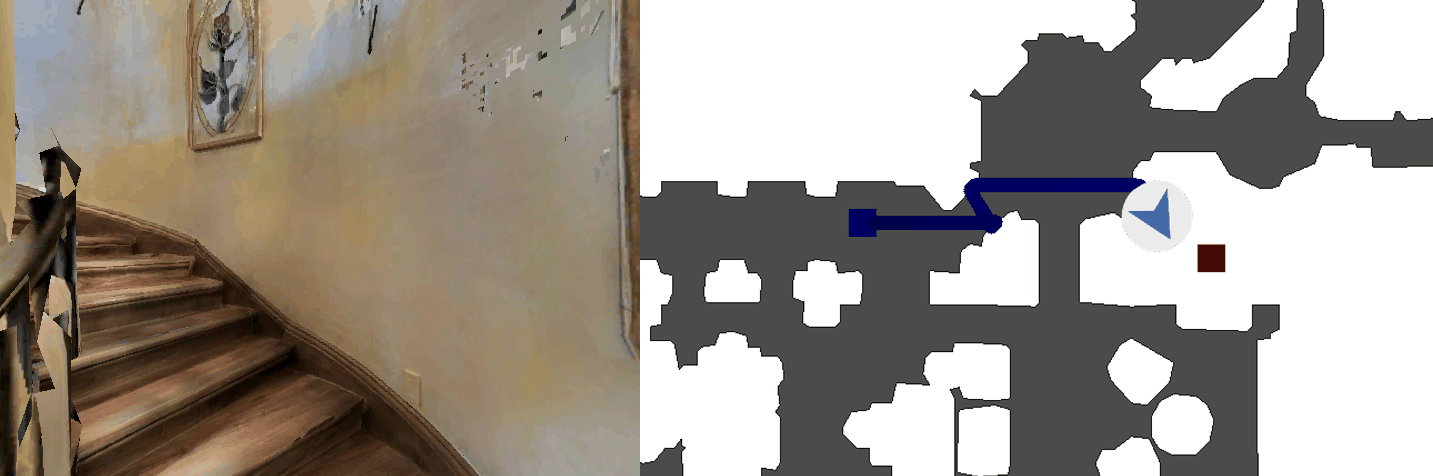}
\end{tabular}
\caption{Navigation trajectories across different scenes showcasing zero-shot perform with both GPT-4.1 and 5.}
\label{fig:simulation_qualitative_results}
\end{figure*}

Furthermore, we demonstrate that \OURS \ is \textbf{model-agnostic}---its performance remains consistent regardless of the foundation model used (in this case GPT-4.1 or GPT-5). \Cref{fig:simulation_qualitative_results} illustrates two sample Habitat episodes, displaying the RGB image and the top-down scene graph produced by our navigation framework.

\subsection{Evaluation of Cache-Assisted Navigation}

\begin{figure}[t]
\setlength\tabcolsep{0pt}
\adjustboxset{width=\linewidth,valign=c}
\centering
\begin{tabular}{c  c c}
    & \OURS\ w/o Caching & \OURS\\\

    \rotatebox[origin=c]{90}{Episode 1475}\hspace{6pt} &
    \includegraphics[width=0.45\linewidth]{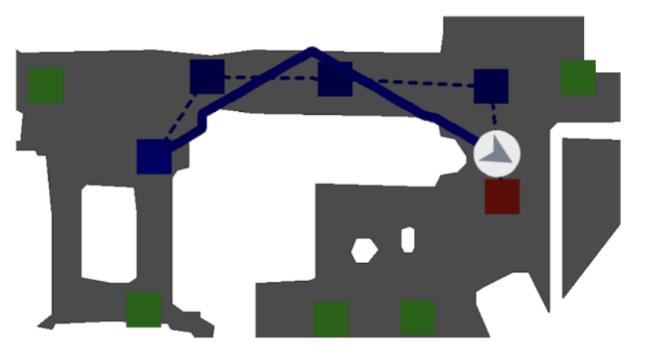} &
    \includegraphics[width=0.45\linewidth]{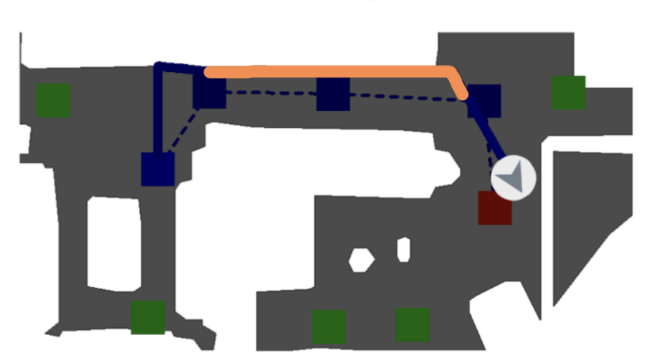} \\

    \rotatebox[origin=c]{90}{Episode 1691} &
    \includegraphics[width=0.45\linewidth]{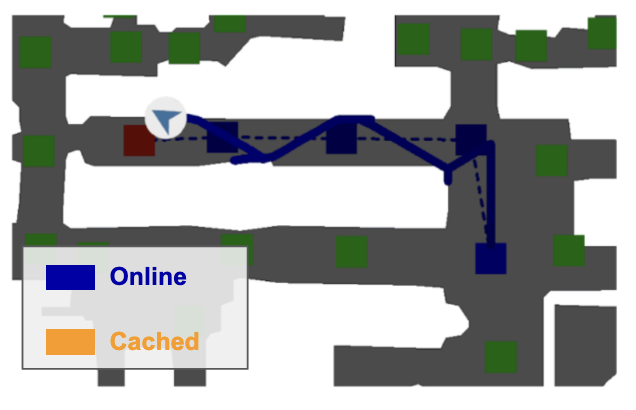} &
    \includegraphics[width=0.45\linewidth]{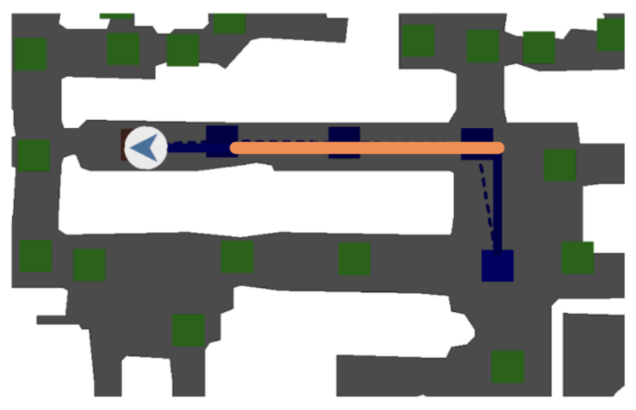}
\end{tabular}
\caption{For two selected episodes in the same scene, we compare paths taken by the VLM with and without cached scene graph edges (orange).}
\label{fig: scene_graph_caching_results}
\end{figure}

To evaluate the effectiveness of our caching framework, we run multiple navigation episodes in a selected scene from the VLN-CE R2R Val-Unseen dataset, after the scene has been explored and the scene graph has been constructed with previously validated paths stored in the location-level cache. We compare the VLM in the cases where it navigates with and without access to cached data in the scene graph. The paths taken are illustrated in \cref{fig: scene_graph_caching_results}, where the path segments highlighted in yellow correspond to subtask-level cache-enabled navigation.

The results are given in \cref{tab: vlmcachingresults}, which shows that our hierarchical framework of cache-enabled navigation allows for consistent reduction in both the total number of VLM calls and average time per iteration, even when an overall task-level trajectory is not present. Implementing caching in our approach allows us to \textbf{reduce the number of VLM calls by up to 78.6\%} and \textbf{the time elapsed by up to 78.8\%} over each episode.



\begin{table*}[t]
    \centering
    \caption{Results from evaluating on different episodes using partially cached scene graph paths in the same scene. We compare \OURS's performance with and without caching across several metrics, including the total number of VLM calls per episode, the average time per step, the total time for episode completion, and the total cost for VLM queries per episode.}
    \vspace{-4pt}
    \resizebox{\textwidth}{!}{
    \begin{tabular}{c|cc|cc|cc|cc} \toprule
         \multirow{2}{*}{Episode} & \multicolumn{2}{c|}{VLM Calls} & \multicolumn{2}{c|}{Avg. Time per Step (sec)} & \multicolumn{2}{c|}{Total Time (sec)} & \multicolumn{2}{c}{Cost (\$)}  \\ 
         & \OURS \ w/o Caching & \OURS & \OURS \ w/o Caching & \OURS & \OURS \ w/o Caching & \OURS & \OURS \ w/o Caching & \OURS \\ \midrule
         1475 & 47 & 32 & 2.161 & 1.352 & 101.551 & 72.996 & 0.200 & 0.136\\
         1691 & 84 & 18 & 1.995 & 0.888 & 167.605 & 35.500 & 0.357 & 0.077\\
         \bottomrule
    \end{tabular}
    }
    \label{tab: vlmcachingresults}
\end{table*}

\begin{figure*}[t]
    \centering
    \includegraphics[width=0.49\linewidth]{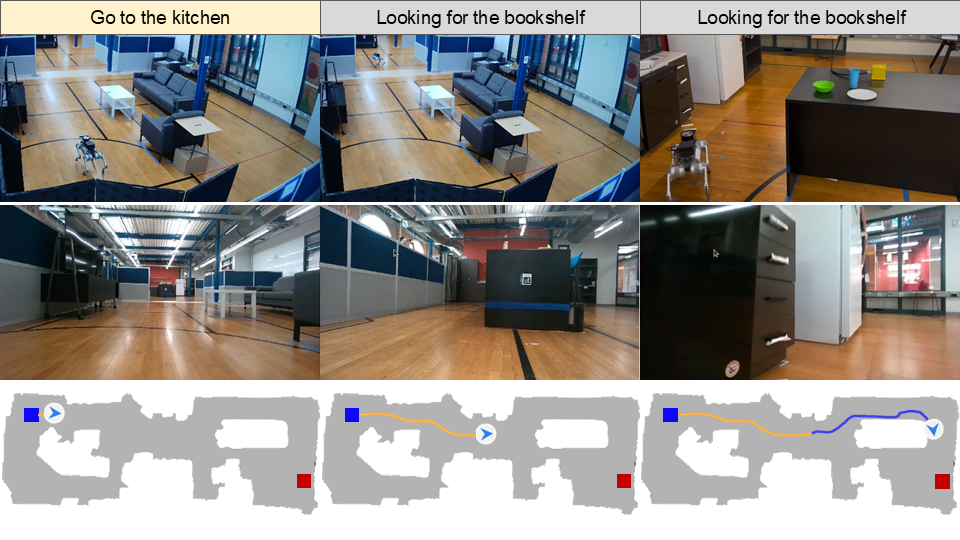}\hspace{-4pt}
    \includegraphics[width=0.49\linewidth]{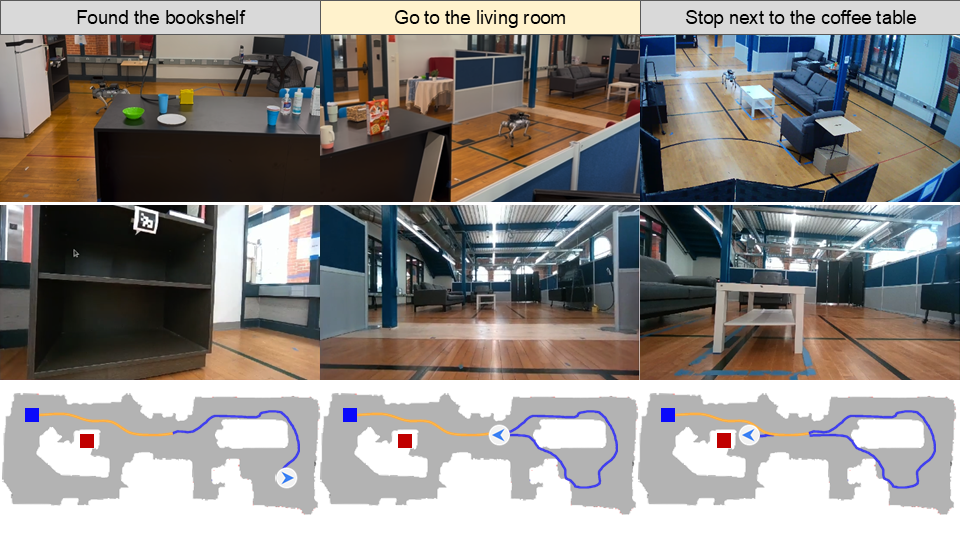}
    \caption{A demonstration of the task ``go to the bookshelf to find a book and return to the coffee table in the living room.'' We show the frames of a third-person-view recording in the top row and a first-person-view recording from the robot's front camera in the bottom row. We label the sub-task at the top of each frame, where the cached sub-tasks are in a yellow. The third row shows the scene graph with incremental trajectories corresponding to each frame.}
    \label{fig: deployment-demo}
\end{figure*}


\section{REAL ROBOT DEMONSTRATION}

In this section, we examine \OURS \ in real-world navigation tasks to provide a step-by-step illustration and demonstrate its capability in the physical world. 

\paragraph{Experimental Setup}

We deploy a quadruped robot, \texttt{Unitree Go2} (as shown at the top of Fig.~\ref{fig: pipeline}), in an apartment consisting of a living room, main entrance, and kitchen. In the exploration phase, we integrate an \texttt{Intel RealSense D456 RGB} depth camera into our VLM-embedded framework, allowing pixel-level detections to project into the map and label rooms, obstacles, and task-relevant objects. We first construct a \emph{scene graph} during exploration guided by the VLM along with odometry information~\cite{mur2015orb}. During execution, we utilize onboard odometry \cite{shan2020lio} to locate the robot in the map in real time.


\begin{figure}[t]
    \centering
    \includegraphics[width=0.85\linewidth]{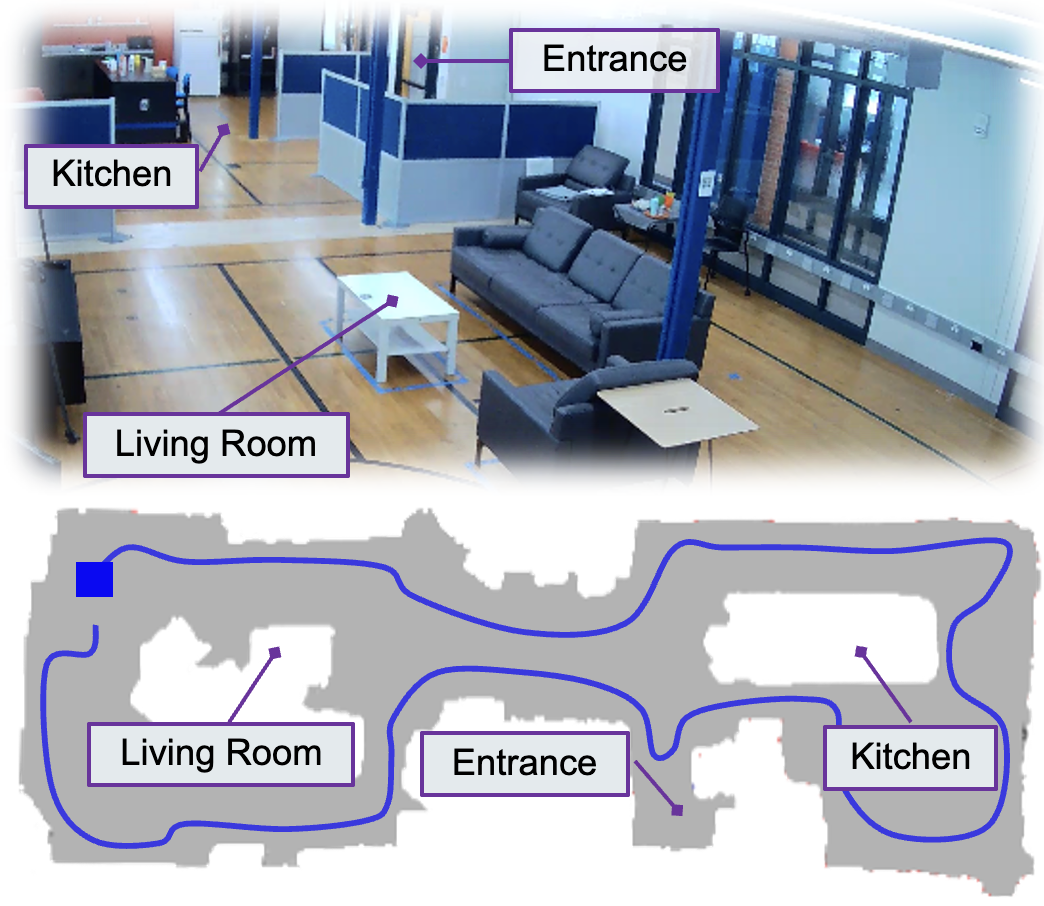}
    \caption{We demonstrate that in an unseen apartment (top) we can construct an associated scene graph. We also show the robot's exploration trajectory (bottom).}
    \label{fig: exploration-demo}
\end{figure}

\paragraph{Exploration} The \texttt{Unitree Go2} traverses the apartment to construct a scene graph of the environment with a single round of exploration, without any revisits, and completes in under ten minutes. Over the course of this run, the robot covers approximately 30m of path length to explore a 30m$^2$ apartment space that includes the living room, entry room, and kitchen. We show the constructed scene graph in Fig.~\ref{fig: exploration-demo}, showing that a single, lightweight exploration pass provides sufficient coverage for deployment.

\begin{figure}[t]
    \centering
    \includegraphics[width=0.9\linewidth]{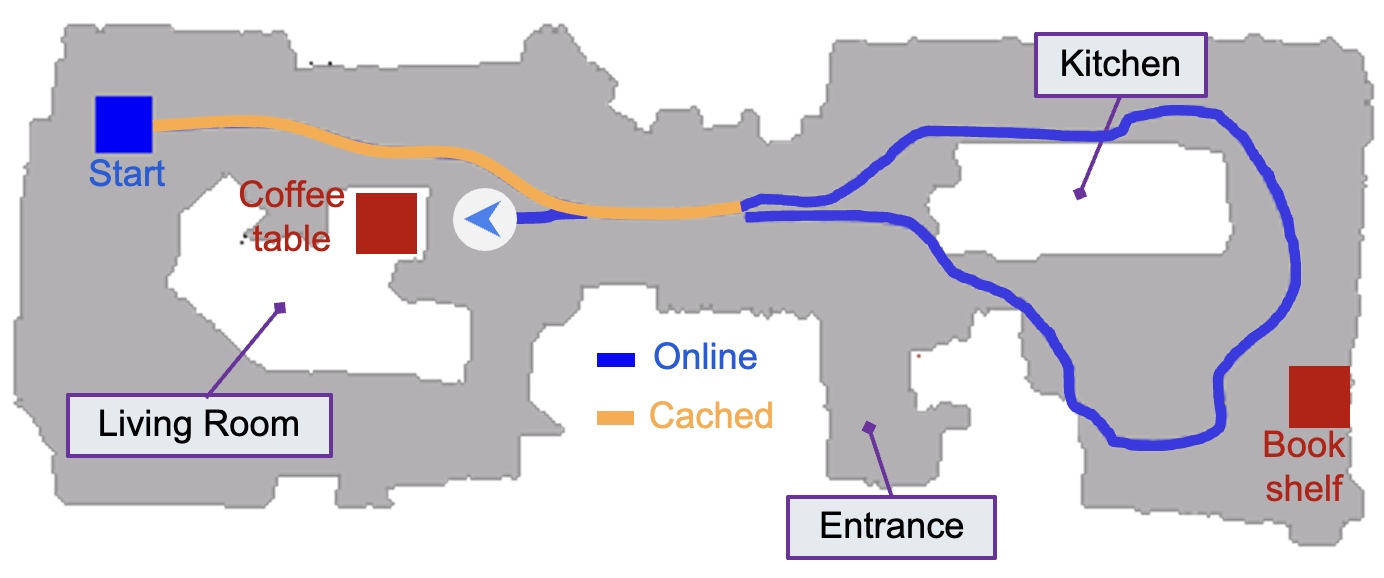}
    \caption{The agent's trajectory is retrieved from subtasks and access to cached trajectories enables efficient deployment.}
    \label{fig: topdownmap-demo}
\end{figure}

\paragraph{Deployment} We assign the robot a task ``Go to the bookshelf to find a book and then return to the coffee table in the living room.'' The planner decomposes this high-level goal into a sequence of subtasks, as illustrated in Fig.~\ref{fig: deployment-demo}. Notably, two subtasks---``go to the living room'' and ``go to the kitchen''---are directly retrieved from the cache, enabling reuse of previously verified trajectories without re-planning. This cache-enabled execution significantly reduces computation and ensures consistency with past safe behaviors. The robot completes the task in approximately three minutes, without any revisit or redundant exploration. We show the corresponding top-down trajectory in Fig.~\ref{fig: topdownmap-demo}, demonstrating how our hierarchical caching achieves efficient and safe execution in a realistic environment.

\section{CONCLUSIONS}

\textbf{Summary: } We develop \OURS, a framework that integrates VLM-guided exploration, neurosymbolic navigation, and hierarchical caching to enable rapid adaptation in unseen environments. By combining rapid exploration for scene graph construction, symbolic reasoning for task decomposition, and cache-enabled execution, \OURS \ addresses inefficiencies and poor generalization in prior robot navigation approaches.
Experiments in both simulation and real-world settings validate these contributions: \OURS \ achieves state-of-the-art zero-shot performance on R2R and RxR benchmarks, reducing navigation error and improving success rate while requiring fewer input modalities. Hierarchical caching further reduces VLM calls and execution time, and our quadruped robot demonstration confirms that these benefits extend to physical deployment. 

\textbf{Future Directions: } We aim to extend \OURS \ to handle out-of-domain environments that differ significantly from indoor environments, e.g. outdoor off-road settings. In addition, we aim to enable navigation under incomplete or partially observed scene graphs. Real deployments often encounter occlusions, dynamic objects, or missing information which requires planners to reason under uncertainty while maintaining safety and efficiency. Addressing these challenges will enable our framework to move beyond existing benchmarks, allowing for reliable operation in complex, dynamic, and open-world environments.

\addtolength{\textheight}{-12cm}   









\bibliographystyle{IEEEtran}
\bibliography{refs}

\end{document}